\documentclass{trbunofficial}

 \usepackage{booktabs} 
 \usepackage{graphicx}
 \usepackage{amsfonts}
 \usepackage{multirow}
 \usepackage{amsmath} 
 \usepackage{array}
\newcommand{\std}[1]{\scriptsize{$\pm$#1}}
\usepackage{amsmath}
\usepackage{amssymb}
\usepackage{caption}
\usepackage{graphicx}
\usepackage{subcaption}
\usepackage[hidelinks]{hyperref}
\usepackage[mathcal]{eucal}

\AuthorHeaders{He, Li, Yuan and Zhou}
\title{A Simple Framework Towards Vision-based Traffic
Signal Control with Microscopic Simulation}

\author{%
  \textbf{Pan He}\\
  Department of Computer Science and Software Engineering\\
  Auburn University, Auburn, AL 36849, USA\\
  pan.he@auburn.edu\\
  \hfill\break
  \textbf{Quanyi Li}\\
  Senior Applied Scientist \\
  Oxa,
Oxford Business Park North
Oxford, OX4 2FL, UK \\
  quanyili0057@gmail.com\\
   \hfill\break
  \textbf{Xiaoyong (Brian) Yuan}\\
 Department of Electrical and Computer Engineering\\
  Clemson University, Clemson, SC 29634, USA\\
  xiaoyon@clemson.edu \\
  \hfill\break
  \textbf{Bolei Zhou}\\
  Department of Computer Science\\
  University of California, Los Angeles (UCLA), Los Angeles, CA 90095, USA\\
  bolei@cs.ucla.edu
}

\begin{document}
\maketitle

\section{Abstract}

Traffic signal control (TSC) is crucial for reducing traffic congestion leading to smoother traffic flow, reduced idle time, and mitigated CO2 emissions. In this paper, we explore the computer vision approach for TSC that modulates on-road traffic flows through visual observation. Unlike traditional feature-based approaches, vision-based methods depend much less on heuristics and predefined features, bringing promising potentials for end-to-end learning and optimization of traffic signals. Thus, we introduce a simple traffic simulation framework called TrafficDojo towards vision-based TSC and its benchmark by integrating the microscopic traffic flow provided in SUMO into the 3D driving simulator MetaDrive. This proposed framework offers a versatile traffic environment for in-depth analysis and comprehensive evaluation of traffic signal controllers across diverse traffic conditions and scenarios. We establish and compare baseline algorithms including both traditional and Reinforcement Learning (RL) approaches. This work sheds light on the design and development of vision-based TSC approaches and opens up new research opportunities\footnote{All the code and baselines have been made publicly available at \url{https://panhe.org/projects/traffic\_dojo.html}}.

\hfill\break%
\noindent\textit{Keywords}: Traffic Signal Control; Reinforcement Learning;
Traffic Simulation; Deep Learning
\newpage

\section{Introduction}
As per the 2022 INRIX Global Traffic Scorecard annual report \cite{pishue2022inrix}, a typical American driver experienced 51 hours in traffic congestion in 2022, resulting in the monetary loss of approximately \$$869$ in wasted time. A key strategy for mitigating traffic congestion is to develop new Traffic Signal Control (TSC) algorithms that can effectively coordinate traffic movements at intersections.

Traffic signals are typically controlled by fixed-time, actuated, or adaptive methods. 
Generally, these traditional methods must be specifically configured or tuned by experts for regions with various traffic conditions, which is labor-intensive and limits its large-scale application. Recently, Reinforcement Learning (RL) has been investigated as a method to optimize adaptive TSC policies for intersections. By learning from scratch through interaction with dynamic traffic environments, RL has demonstrated superior performance compared to traditional approaches~\cite{mei2022libsignal,tran2023revisiting,islam2023effect}.
However, the current study of RL-based methods remains insufficient, as they operate under the strong assumption that all high-level features and observations---such as queue length, waiting time, volume, delay, speed, vehicle positions, phase duration, and congestion---are readily available~\cite{wei2019survey}. 
These high-level heuristics are often difficult to extract from raw sensor inputs.
Connected and Autonomous Vehicles (CAVs), leveraging Vehicle-to-Vehicle (V2V) and Vehicle-to-Infrastructure (V2I) communication, can enhance TSC by providing heuristic information on the speed and location of surrounding vehicles, thereby reducing reliance on camera-based systems. However, they~\cite{chen2021distributed,song2022impacts} are unsuitable for environments where human-driven vehicles dominate. Such scenarios necessitate cameras for sensing and perception algorithms to detect and respond to these vehicles accurately.  It is fair to assert that RL-based TSC needs further exploration to determine if its policy remains robust when operating with noisy high-level features derived from cameras.

Additionally, manually engineered high-level features may not comprehensively represent traffic scenarios or align with specific reward designs, hindering RL-based approaches from identifying the optimal solution.
In contrast, the end-to-end policy that directly takes raw images as input has the potential to learn a better feature extractor and boost the performance of the TSC policy.
Unfortunately, there is a limited body of work exploring vision-based TSC methods, and many existing studies focus on training TSC policies using simplified or toy top-down snapshots~\cite{van2016coordinated,tran2023revisiting}. Moreover, the 2D nature of the simulation limits the ability to study multiview camera setups in a 3D environment. This paper aims to explore end-to-end TSC policy with visual inputs.

\begin{figure*}[!t]
    \centering
    \includegraphics[width=0.9\textwidth]{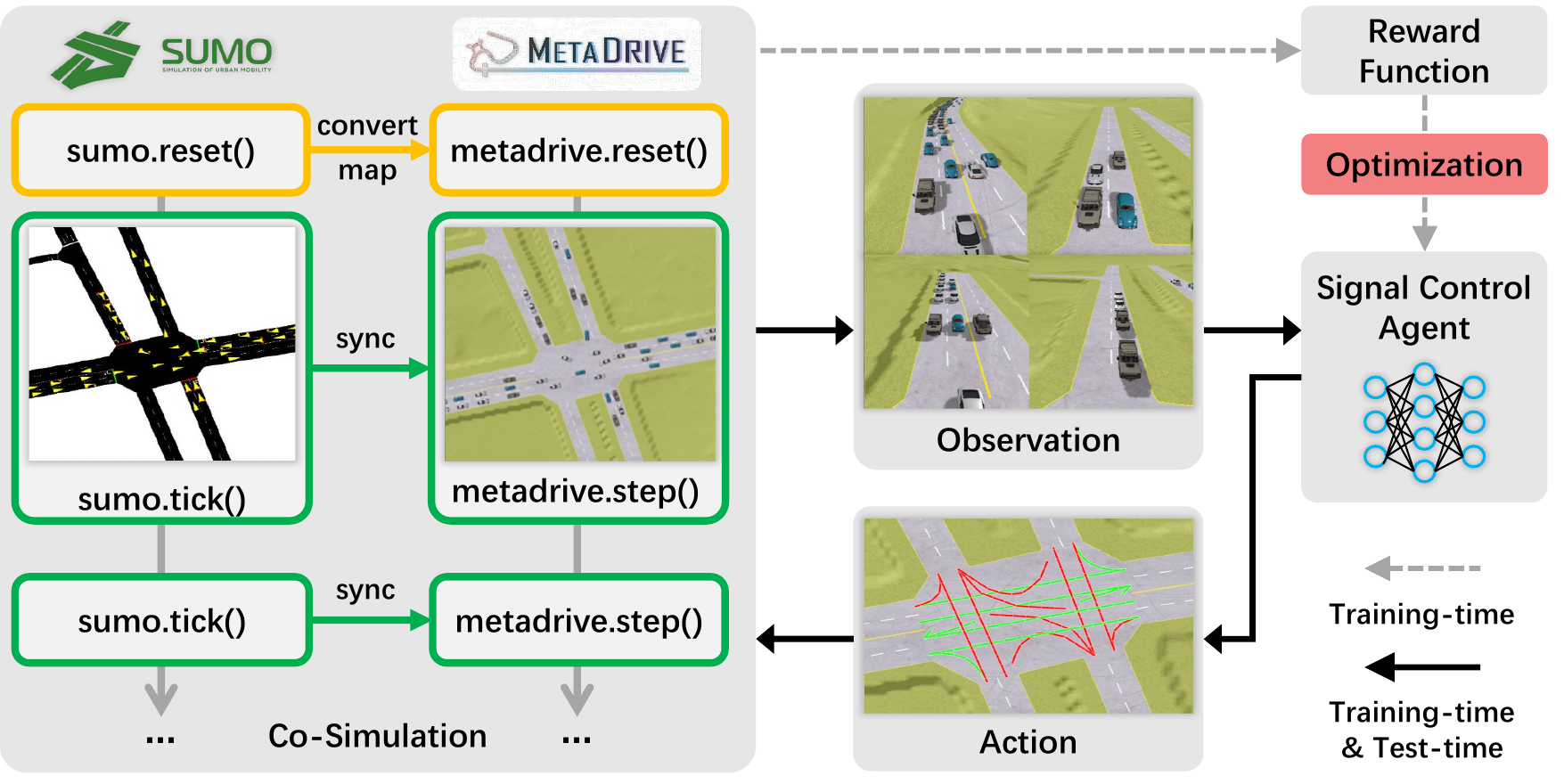}
    \caption{Overview of TrafficDojo. TrafficDojo supports generating rich 3D visual scenarios from SUMO maps, leveraged by a visual rendering engine of MetaDrive. At each time step, TrafficDojo implements a synchronization mechanism to direct the synchronous creation, updating, and removal of vehicles and pedestrians between SUMO and MetaDrive. TrafficDojo thus provides a Gym interactive environment tailored for traffic signal control, equipped with the capability to capture visual data from sensors like RGB cameras and LIDARs positioned at a traffic intersection. Additionally, it directly connects to the common RL training platforms such as Stable-Baseline3~\cite{raffin2021stable} and RLLib~\cite{liang2018rllib} where a wide range of RL algorithms can be evaluated. 
}

    \label{fig:system}
\end{figure*}

Although these aforementioned questions are essential for accelerating the research and deployment of vision-based TSC methods, we cannot tackle them using existing research infrastructures or traffic simulators. 
Popular traffic simulators, such as Flow~\cite{wu2017flow}, Simulation of Urban MObility (SUMO)~\cite{lopez2018microscopic}, CityFlow~\cite{zhang2019cityflow}, RESCO~\cite{ault2021reinforcement}, and CBLab~\cite{liang2023cblab}, have been introduced to simulate diverse traffic scenarios.
However, none of them support sensor simulation and thus can not serve as the test-bed to investigate the impact of the high-level feature estimation for vision-based TSC methods.
A potential solution is to use CARLA~\cite{dosovitskiy2017carla} with the SUMO bridge where the images are rendered by CARLA and traffic flows are generated by SUMO.
Nevertheless, this solution is not efficient because CARLA is too slow to render images to train RL algorithms.
As a result, it may take several days or even weeks to collect enough samples for training a TSC policy.

In this work, we develop an open-source traffic simulation framework, TrafficDojo, for fast-prototyping and comparing TSC algorithms.
Our TrafficDojo seamlessly integrates MetaDrive~\cite{li2022metadrive} and SUMO simulators into a unified simulation system where traffic is controlled by SUMO while MetaDrive serves as a rendering engine that provides multi-view visual observations.
Compared with CARLA~\cite{carla2018release080}, which achieves roughly 45~FPS with a single camera on low-quality settings in a moderately complex scene, MetaDrive provides a more favorable balance between rendering fidelity and simulation efficiency. Notably, MetaDrive can sustain up to 300~FPS on a standard PC while maintaining high-quality visuals and a lightweight footprint of only about 100~MB per instance~\cite{li2022metadrive}.
Also, it is lightweight and can be easily parallelized.
To facilitate the applicability of standard RL libraries, we follow the standard OpenAI Gym interface and provide a TSC environment that enables capturing visual data from RGB cameras.
The system is programmed in Python and will be open-sourced.
In summary, the main contributions of the paper are:
\begin{itemize}
    \item We develop TrafficDojo, a traffic signal simulation framework that bridges MetaDrive with microscopic simulation and facilitates the research of vision-based TSC policies. The modular-based design enables customizable configurations of visual sensors and map topologies.
     \item We benchmark various vision-based adaptive TSC approaches across two standard traffic scenarios and offer a comprehensive set of evaluation metrics reflecting quality of service and environmental impact.
    \item We investigate the vision-based adaptive TSC approaches and provide valuable insights into their model design and development and demonstrate that using a pre-trained foundational model as a feature extractor for vision-based TSC approaches is highly effective, surpassing traditional feature-based methods.
\end{itemize}

Experimental results highlight the promising potential of end-to-end learning and control for vision-based TSC in simulated environments. We hope this approach will open up new research opportunities for both the reinforcement learning and transportation communities.

\section{Related Work}
In this section, we delve into recent research that aligns with our work, specifically focusing on traffic signal control, simulation environments, and road network generation and descriptions.

\subsection{Traffic Signal Control} 

 TSC strategies, categorized into fixed-time, actuated, and adaptive control strategies~\cite{wei2019survey}, aim to optimize signal timings for efficient traffic flow. Fixed-time methods~\cite{miller1963settings} rely on historical data to establish predetermined signal control sequences, assuming stable traffic conditions over a specific period, typically the past five minutes. 
Actuated signal methods adapt to real-time traffic data from sensors like pressure plates and loop detectors, via determining phase and duration based on current demand and follow predefined logical rules. 
Adaptive methods~\cite{hunt1981scoot} optimize signal timing and excel at automatically adjusting to unexpected traffic changes by  monitoring and addressing performance gaps.


Recent advancements in TSC leverage RL methods that adapt to real-time traffic conditions through trial-and-error~\cite{van2016coordinated,tran2023revisiting,ducrocq2023deep}, unlike non-RL approaches relying on specific assumptions. Traditional RL methods often use low-dimensional, hand-crafted features as state representations, limiting performance~\cite{wei2019survey}. RL-based methods can generally be categorized by reward, state, and action designs. Typical rewards include queue length, waiting time, change of delay, speed, number of stops, throughput, frequency of change, accident avoidance, and pressure~\cite{wei2019survey}.
Many state representations have been employed to describe the traffic conditions, such as queue length, waiting time, volume, delay, speed, positions of vehicles, phase duration, and congestion. In terms of action designs, several common choices exist, such as establishing the current phase duration~\cite{aslani2017adaptive}, determining the ratio of the phase duration relative to a predetermined total cycle duration~\cite{casas2017deep}, transitioning to the next phase in a predetermined cyclic phase sequence~\cite{pham2013learning,mannion2016experimental,van2016coordinated}, and choosing the phase to transition to from a set of available phases~\cite{arel2010reinforcement,zheng2019diagnosing}. 
We refer the readers to \cite{arulkumaran2017brief} for a comprehensive overview of RL algorithms. However, recent deep RLs have made processing high-dimensional input data like images feasible, offering new TSC possibilities. This paper uses visual data to compare RL-based TSC algorithms against traditional controllers. 

\subsection{Simulation Environments} 
Traffic simulation includes three main categories: macroscopic, microscopic, and mesoscopic. Macroscopic approaches describe traffic flow at an aggregate level, focusing on parameters like traffic speed, flow, and density and large-scale road simulation. 
Microscopic approaches model individual road users, such as vehicles, and their interactions with the environment, including acceleration, braking, lane-changing, and traffic signals. Traffic3D~\cite{garg2019traffic3d} leverages Unity's graphical and physics engines but is limited to a narrow range of road traffic scenarios. Sumounity~\cite{pechinger2024sumonity}, a concurrent work, integrates SUMO with Unity to enhance traffic simulation capabilities. SUMO, a popular microscopic simulator, supports 2D simulations of vehicles, public transport, and pedestrians, with extensions like CityFlow~\cite{zhang2019cityflow} and FLOW~\cite{wu2017flow} enabling diverse traffic scenarios. SUMO also allows co-simulation with 3D platforms like CARLA~\cite{dosovitskiy2017carla} and Unity 3D. We focus on SUMO for its ability to simulate long-term traffic scenarios, despite its limitation in generating synthetic sensor data. 
Mesoscopic approaches are intermediate approaches bridging macroscopic and microscopic. We refer the readers to~\cite{boxill2000evaluation} for a comprehensive review. In this paper, we aim at developing the benchmark and library for vision-based TSC.

\subsection{Road Network Generation and Descriptions} 
Effectively acquiring and modeling road networks is a crucial yet challenging task in traffic simulation. While digital representations of real-world road networks are increasingly available,  their direct integration in simulations often require multi-step procedure and refinements, requiring domain expertise for seamless incorporation. CityEngine~\cite{parish2001procedural} proposes a procedural modeling approach in the generation of 3D urban environments. MITSIM describes the semantics of a road network using terms \textit{node},  \textit{link},  \textit{segment}, and  \textit{lane}. VISSIM~\cite{fellendorf1994vissim} introduces  \textit{link} and  \textit{connector} to represent the topology of a road network. OpenSCENARIO defines an open standard for describing the dynamic content of driving scenarios. OpenDRIVE~\cite{dupuis2010opendrive} standardizes the logical road description between different driving simulators. OpenCRG~\cite{asam2023openCRG} is developed to describe 3D geometries such as inclination, pitch, and yaw angles of the roads. CommonRoad~\cite{althoff2017commonroad} leverages lanelets~\cite{bender2014lanelets} to describe road networks. SCENIC~\cite{fremont2019scenic} introduces a new probabilistic programming language for describing driving scenarios. 
Using procedural generation and importing real-world data, MetaDrive~\cite{li2022metadrive} can create an infinite number of scenarios. A comprehensive surveys is suggested in~\cite{ma2021traffic}. In this paper, we utilize MetaDrive's scenario definitions to represent traffic scenarios. Additionally, we integrate MetaDrive as the rendering engine due to its lightweight nature and efficiency.

\section{Preliminaries}
In this section, we provide the preliminary knowledge about traffic signal control tasks. 

\subsection{Terminologies}
We use a standard intersection with four approaches as an illustrative example to demonstrate the key concepts. These concepts can be extended to intersections with various topological structures, such as varying numbers of approaches and skewed intersections.

\begin{itemize}
    \item \textbf{Approaches} refer to the road segments or lanes directing traffic into or out of an intersection, specifying the distinct directions from which road users approach or depart. In a typical intersection, there are four incoming approaches, namely North, South, West, and East. Accordingly, there are outgoing approaches on which vehicles can leave the intersections.
    \item \textbf{Traffic Movement} describes the various ways in which road users navigate through an intersection. Each movement corresponds to a specific path or direction a vehicle, pedestrian, or cyclist can take within the intersection. Some examples of traffic movements include a left turn, right turn, through movement, or U-turn.
    \item \textbf{Traffic Signals}, commonly known as traffic lights, serves as mechanisms for regulating traffic flows at intersections. They determine when specific traffic movements are permitted or required to stop, guaranteeing the safe and efficient movement of vehicles and pedestrians. The standard components include red, green, and yellow lights.
    \item \textbf{Phases} describes the distinct operations of a signalized intersection, equipped with multiple lanes for incoming and outgoing roads. Each phase represents a specific combination of traffic movements. Phases are identified as ``in conflict'' when simultaneous assignment is not feasible, often due to intersecting traffic movements.   A signal controller is responsible for assigning non-conflicting phase sequences to optimize the long-term objective function. We direct readers to~\cite{wei2019survey} for illustrative examples depicting both non-conflicting and conflicting phases, and commonly used phases.
    \item \textbf{Green Time}, also referred to as the duration of the green phase, designates the period during which traffic can safely pass through the intersection. The actual duration of this phase is determined by the traffic conditions in the studied area and the space available for vehicles waiting in the queues. To ensure the safety of vehicles and pedestrians, it is critical to minimize travel time and simultaneously prevent any skipping of the green phase. Therefore, there is often a minimum green time associated to an intersection.

\end{itemize}



\subsection{Traffic Signal Control as Markov Decision
Process}

Reinforcement Learning trains an agent how to act to maximize a given numerical reward function. In our TSC setting, RL will train the agent to map situations to a favorable action to control the traffic light. It is modeled as a Markov Decision Process (MDP) defined as $\mathcal{M} = \{\mathcal{S},\mathcal{A}, \mathcal{P}_a, \mathcal{R}, \gamma \}$, where $\mathcal{S}$ is the \textit{state space}, $\mathcal{A}$ is the \textit{action space}, $\mathcal{P}_a(s, s') = \Pr (s_{t+1} = s' \mid s_t=s, a_t = a) $ defines the probability of transitioning from state $s$ at time $t$ to state $s'$ at time $t+1$ given action $a$. $\mathcal{R}: \mathcal{S} \times \mathcal{A} \rightarrow \mathbb{R}$ denotes the \textit{reward function}, and $\gamma \in [0,1]$ is a discount factor. At each timestep $t$, an agent relies on a state $s_t \in \mathcal{S}$ to take an action $a_t \in \mathcal{A}$, which is driven by a policy $\pi : \mathcal{S} \rightarrow \mathcal{A}$. The next state $s_{t+1}$ is then determined by the transition probability $\Pr(s_{t+1} \mid s_t, a_t)$ while the agent received a reward $r(s_t, a_t) \in \mathcal{R}$. To learn the decision-making process, we leverage the accumulated award $G_T = \sum_{t = 0}^T \gamma^t r(s_t, a_t)$ to find the optimal policy $\pi^*$ such that $G_T$ is maximized. Popular RL approaches generally include value-based, policy-gradient, and actor-critic approaches. 

The TSC can be formulated as an MDP using RL approaches. The \textit{State Space $\mathcal{S}$} describes the state of incoming traffic and current phase information. Specifically, the state of incoming traffic is determined by the sensing capability and ranges that are assumed in the environment. Our TSC agent representing the state via directly mapping RGB images (with a resolution of $256 \times 256$) to actions for controlling the traffic signal. This leads to an end-to-end learning, that might eliminate the need for pre-specification of traffic environment features, such as vehicle density and queue length.
At each timestep, the agent chooses an action, i.e., changing to a signaling phase for assigning right-of-passage (green light), from  the \textit{Action Space $\mathcal{A}$}. Figure~\ref{fig:sensor} illustrates a visual example of observations and actions. A mandatory yellow phase is enforced for a specified duration if the chosen phase differs from the current phase, so as to obey the the real-world phase transition rules. 
The \textit{Transition Function $\mathcal{P}_a$} is determined by traffic progression models for simulated environments, i.e., MetaDrive and SUMO. The \textit{Reward Function $\mathcal{R}$} reflects the objectives or interests of traffic agents. In this paper, we focus on optimizing traffic throughput and reducing travel time or delay of vehicles. Therefore, we adopt the popular diff waiting time as the reward, which measures the change of waiting time of vehicles~\cite{sumorl}.

\section{Overview of TrafficDojo}

In this section, we introduce an open-source traffic simulator, called TrafficDojo for developing and benchmarking vision-based adaptive TSC algorithms. We describe the key designs of TraffiDojo and introduce its executing logic for TSC tasks.



\subsection{Overview}\label{sec:simulation}

Both SUMO and MetaDrive have been chosen as key simulation tools within our TrafficDojo framework. MetaDrive, specifically, stands out as a lightweight and flexible framework built on Panda3D. It offers a precise physics simulation and supports a range of sensory inputs. 
Notable is MetaDrive's remarkable balance between visual rendering and physical simulation, achieving up to 300 FPS on a standard PC with a single instance of 100 MB in size~\cite{li2022metadrive}.
TrafficDojo seamlessly integrates MetaDrive, leveraging its distinctive features while maintaining comparable simulation speed. However, MetaDrive exhibits limitations in its ability to comprehensively analyze diverse traffic flow conditions, primarily due to its original design focusing on short-term self-driving scenarios from the perspective of the ego vehicle. Hence, we have seamlessly integrated SUMO~\cite{lopez2018microscopic}, an open source, highly portable, microscopic traffic simulation platform, into our TrafficDojo framework. This choice is attributed to SUMO's exceptional capability in efficiently managing large-scale networks and accurately simulating realistic traffic flows. Significantly, SUMO utilizes the TraCI (Traffic Control Interface) API for dynamic traffic modeling and quick customization of traffic scenarios.

We make efforts in bridging MetaDrive with SUMO for large-scale traffic simulation. While CARLA offers co-simulation with SUMO, it sees limited adoption in the TSC community due to the intricacy of creating custom maps and using them to run simulations in CARLA. We offer a more practical and efficient choice for researchers to train and test their algorithms and protocols, focusing on the downstream TSC application. Specifically, our research developments equip TrafficDojo with several key features:
\begin{itemize}
    \item a map manager to facilitate the seamless transformation of road network configurations and the support of importing maps and managing traffic flow.
    \item a synchronization manager to direct the synchronous creation, updating, and removal of vehicles and pedestrians between SUMO and MetaDrive.
    \item a traffic signal agent manager takes charge of traffic signal controllers at intersections, responsible for retrieving critical traffic information and making transitions between traffic phases. It also provides an interface to adjust the behaviors and actions of traffic controllers, facilitating the exploration of diverse TSC approaches.
    \item an observer function linked to traffic signals, capturing visual information from a specific viewpoint. This functionality enhances data collection capabilities, providing valuable insights into traffic dynamics and contributing to a more comprehensive understanding of the traffic environment through the utilization of visual data.
\end{itemize}

\subsection{Observing the Traffic Situation}

\begin{figure}[h]
  \centering
  \begin{subfigure}[b]{0.45\textwidth}
    \includegraphics[width=\linewidth]{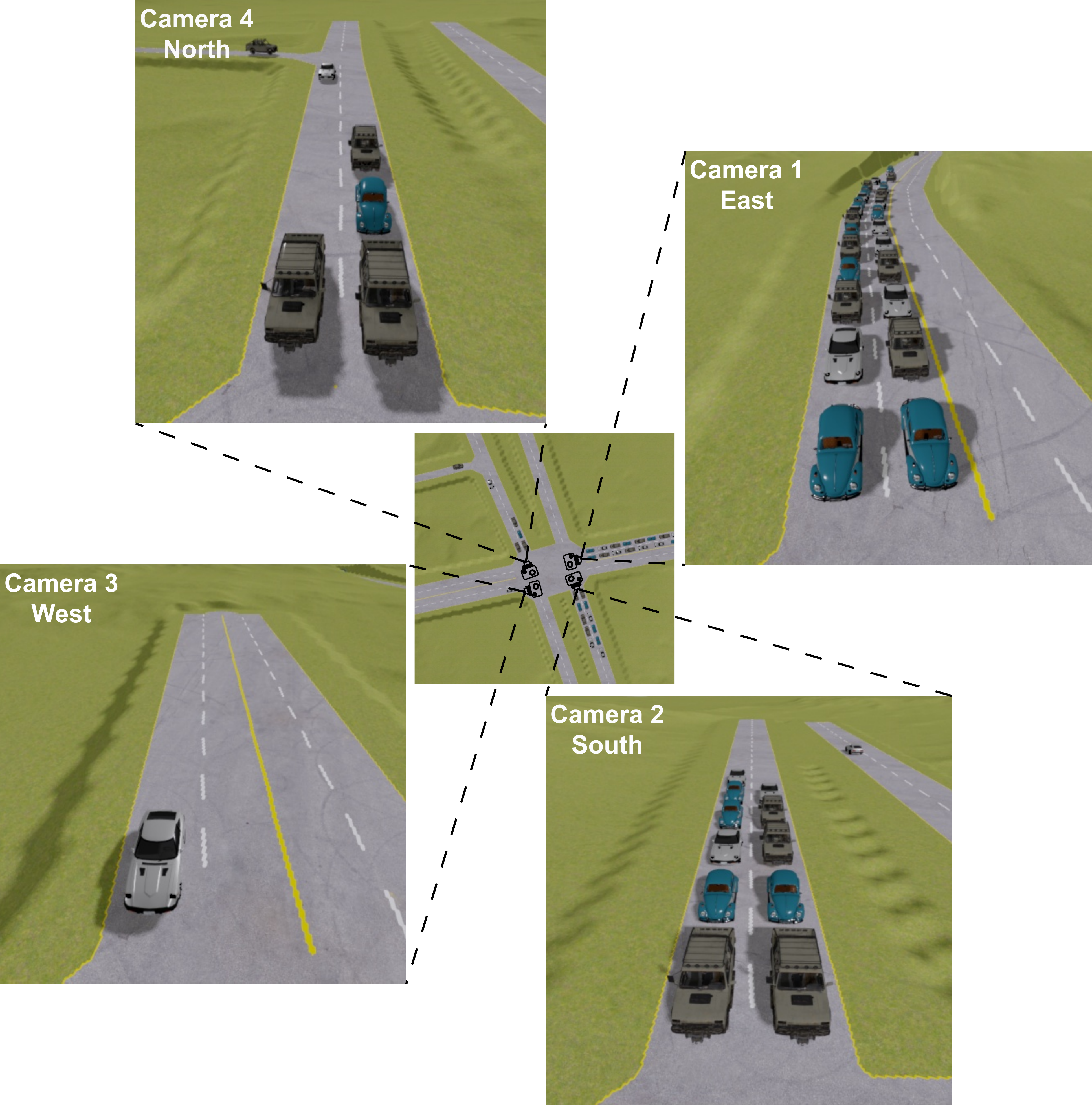}
    
  \end{subfigure}
  \hfill
  \begin{subfigure}[b]{0.45\textwidth}
    \includegraphics[width=\linewidth]{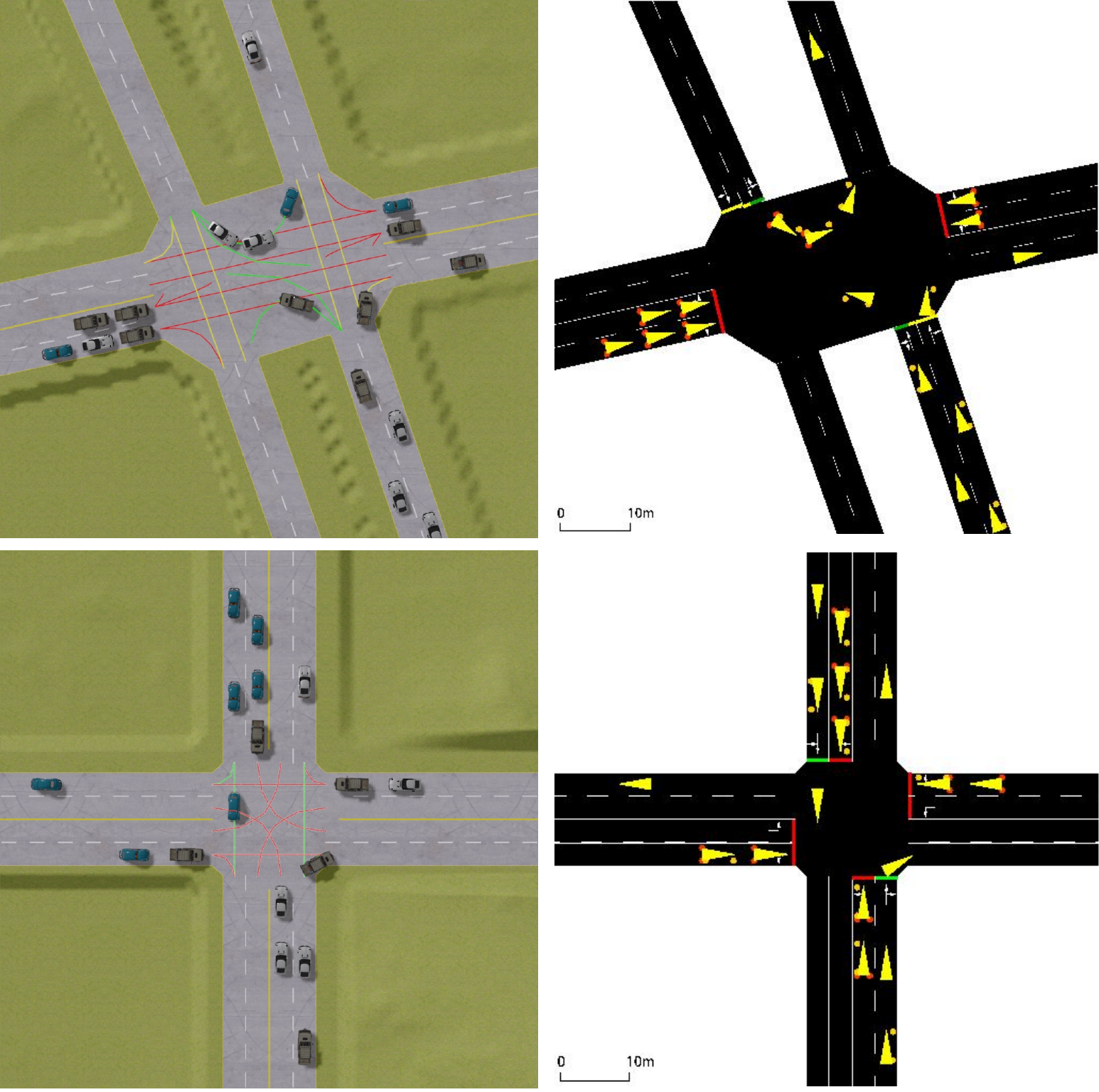}
    
  \end{subfigure}
  \caption{(Left) The multi-view observation setup at an intersection. It involves cameras positioned at a height of 10 meters, each focusing on the respective approaching lanes. The camera height can be adjusted to provide images in different views. (Right) The coordination of traffic signals and participants between MetaDrive and SUMO is achieved through synchronization. The status of traffic lights for specific links is represented by green, yellow, and red lines. The left image is the rendered image from MetaDrive and the right image is the SUMO visualization.}\label{fig:sensor}

\end{figure}

Our framework allows flexible sensor placement (e.g., LIDAR, RGB cameras, depth cameras) and fine-tuning of parameters like field of view and mounting height. It also features an observer function that dynamically adjusts sensors based on traffic signal location and surrounding environment. This enables the strategic placement of multiple cameras, each precisely directed towards the approaching lanes.  By doing so, it effectively captures and monitors traffic from those lanes. The heights of street light poles vary depending on the area they serve, which in turn affects the mounting height of the cameras. According to~\cite{fullerton1990traffic}, cameras are typically mounted at heights ranging from 30 to 60 feet (9 to 18 meters). Therefore, in this paper, we heuristically place each camera at a height of $10$ meters and set its pitch angle to be $-35^{\circ}$. Figure~\ref{fig:sensor} presents exemplary illustrations of the multiview camera setup (left) and the corresponding bird’s-eye view (right).

\subsection{Generation of Road Networks}

We provide the map manager to generate rich 3D scenarios from SUMO maps. It translates the SUMO format for scenario definitions into a unified description, providing precise geometric details for map objects such as driving roads, sidewalks, shoulders, lanes, and crossings. By harnessing OpenStreetMap data, which offers detailed road network models, including layouts, intersections, traffic rules, and geographical features, TrafficDojo enables the design and evaluation of TSC approaches using real-world geographical data. It's crucial to note that in cases where the conversion encounters errors, such as confusing connections or sidewalk overlapping, particularly in large-scale urban networks sourced from OpenStreetMap, manual refinement is still necessary via tools such as SUMO-GUI for customized maps.

\subsection{Synchronization of Traffic Signals and Participants}

Inspired by CARLA co-simulation, we developed a synchronization manager to efficiently handle the dynamic creation and removal of vehicles and pedestrians while seamlessly updating their positions between SUMO and MetaDrive. Furthermore, our framework provides an interface for controlling and synchronizing the status of traffic signals between both. 
A visual example is shown in Figure~\ref{fig:sensor} (right).
The traffic signal manager takes charge of controllers at intersections. At each fixed-time step, the controller selects a non-conflicting phase to receive the right-of-passage (green light). If the selected phases differ from the currently active ones, a mandatory yellow phase is triggered for a predefined duration, such as two seconds. Additionally, a minimum green time is ensured for all phases.

\begin{figure}
    \centering
    \includegraphics[width=0.6\textwidth]{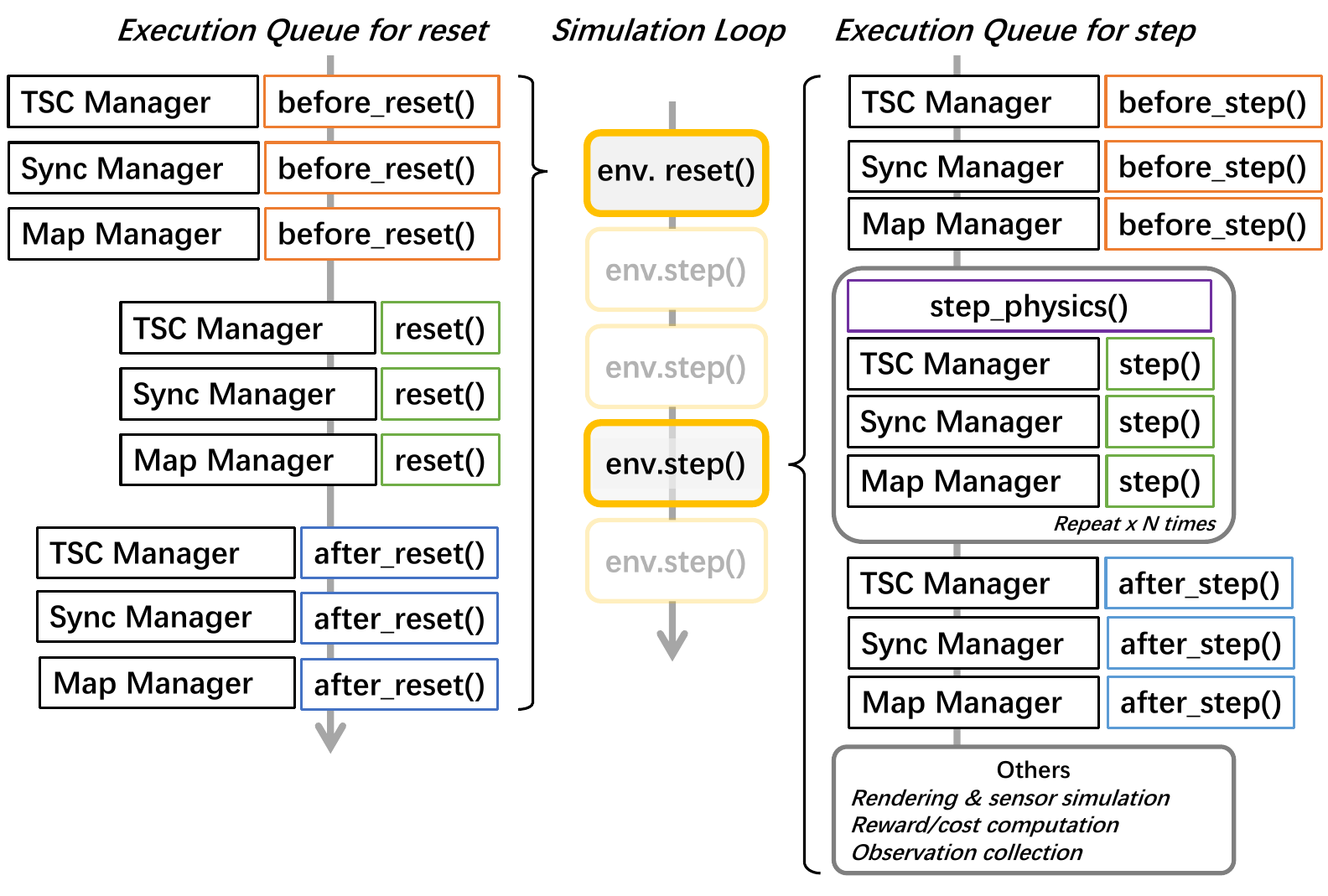}
    \caption{The execution logic of TrafficDojo. Multiple managers are registered and executed in specific orders for managing traffic signal controllers, synchronizations, and maps.}
    \label{fig:execution_queue}
\end{figure}

\subsection{The Execution Logic}

At every simulation step, TrafficDojo oversees the simulator's status, managing static objects like maps and dynamic objects' attributes such as positions, headings, and velocities of vehicles, along with traffic signal status. Figure~\ref{fig:execution_queue} depicts TrafficDojo's execution logic, with several defined key functions for each manager. 

\begin{itemize}
    \item \textit{before\_reset}. It involves the specification of essential preparation functions required before initializing the manager including tasks like cleaning up traffic objects and detaching the map from the scene.
    \item \textit{reset}. This step generates objects according to some predefined rules such as initializing and attaching the map to the scene and initializing/restarting simulators.
    \item \textit{before\_step}. It is used to configure manager actions for tasks such as  communication with SUMO and setting the light status in our specific context.
    \item \textit{step}. This function preprocesses actions for specified agents, e.g., TSC agents,  and performs the underlying physics simulation using the Bullet engine. 
    \item \textit{after\_step}.  It allows post-processing, calculation, and information collection after advancing the simulation.
\end{itemize}

It is important to note that the inclusion of each key step is optional and depends on the specific requirements of a given task. In our framework, managers serve as comprehensive controllers for almost everything, including maps, vehicles, pedestrians, traffic lights, etc.

\section{Experiments}

In this section, we establish different baseline TSC controllers and compare their performance with feature- and vision-based TSC algorithms in different datasets.

\subsection{Benchmark Algorithms}

 \underline{Baseline controllers:} 1) \textit{FixedTime} controls the traffic signal according to a predefined time duration and phase order, whether traffic is present or not.  2) \textit{Self-Organizing Traffic Lights (SOTL)}~\cite{cools2013self},  is a cycle-based dynamic
phase length TSC based on self-organizing principles. 3) \textit{MaxPressure}~\cite{varaiya2013max} is designed to create an acyclic, dynamic phase length TSC, via modeling vehicles in lanes as a substance flowing through a pipe and implements control to maximize the relief of pressure between incoming and outgoing lanes. 

\noindent  \underline{RL controllers:} 1) \textit{Deep Q Network (DQN)}~\cite{mnih2015human} is a powerful algorithm that combines deep neural networks with Q-learning to learn optimal policies in complex environments.
   2) \textit{A2C}, short for Advantage Actor Critic, presents a synchronous variant of the A3C method~\cite{mnih2016asynchronous}. Unlike A3C, A2C is synchronous and deterministic, waiting for each actor to complete its experience segment before updating, subsequently averaging across all actors. 
3) \textit{Proximal Policy Optimization (PPO)}~\cite{schulman2017proximal} integrates concepts from A2C and TRPO.

We explore four distinct model variants. 1) \textit{Feature-based RL approaches (DQN-F, PPO-F, A2C-F)} employ predefined features from SUMO, such as lane density and queue length. 
2) \textit{Vision-based RL approaches with multi-view cameras (DQN-MV, PPO-MV, A2C-MV)}: In contrast to feature-based variants, these models utilize multiple cameras to capture visual information, as shown in Figure~\ref{fig:sensor}. We limit our exploration to straightforward methods for fusing features from multi-view cameras with a direct concatenation operation.
3) \textit{Vision-based RL approaches with a single BEV camera (DQN-BEV, PPO-BEV, A2C-BEV)}: Using a BEV camera to observe the traffic intersection.
4) \textit{Vision-based RL approaches utilizing a single BEV camera and a frozen foundational model for feature extraction (DQN-BEV*, PPO-BEV*, A2C-BEV*)}: Addressing the challenges of learning directly from high-dimensional images~\cite{tran2023revisiting}, these models use a frozen foundational model for feature extraction. Inspired by CLIP~\cite{radford2021learning}, we replace feature extractors in DQN-BEV, PPO-BEV, and A2C-BEV with the pretrained ViT-H-14-quickgelu model from OpenCLIP~\cite{ilharco_gabriel_2021_5143773}. This enhancement builds policies on robust features, resulting in the modified DQN-BEV*, PPO-BEV*, and A2C-BEV*.

\subsection{Evaluation Metrics} 


We evaluate TSC algorithm thoroughly using the following metrics: 1) Average travel time, calculating the average duration spent by each vehicle traveling the network including both waiting and actual travel time; 2) Throughput, normalized on an hourly basis, counts successfully completed trips; 3) Queue length, the sum of vehicles waiting on lanes; 4) Average delay, calculated as $1 - \frac{\sum_{i=1}^n v_i}{n * v_{max}}$, characterizing the difference between mean vehicle speed and maximum speed limit, where $n$ is the number of vehicles on the lane, $v_i$ is the mean speed of vehicle $i$, and $v_{max}$ is the maximum allowed speed; 5) Accumulated waiting time, the sum of vehicles' waiting time across lanes over a specific interval (e.g., $1000$); 6) CO2 emissions, evaluating environmental impact by computing the total emitted CO2 by vehicles.

\subsection{Experiment Setup}


We evaluated two representative traffic scenarios: a synthetic two-way intersection and a real-world TAPAS Cologne intersection~\cite{varschen2006mikroskopische} with calibrated traffic demands. Simulations ran for $3,600$ seconds, with our map manager converting road networks into suitable representations. Key settings included a 10-second minimum green time, a 5-second action interval, and a mandatory 2-second yellow phase before green transitions.  All baseline models were implemented in our TrafficDojo. RGB images from multi-view and BEV sensors are configured with a resolution of $256 \times 356$. Experiments were conducted on a server with two AMD EPYC Milan 7713 CPUs and four NVIDIA RTX A6000 GPUs. Results are based on the best-performing models selected from 200 training epochs, with evaluations averaged over five runs, reporting means and standard deviations. All code and baseline implementations have been publicly released to help readers better understand the details of TrafficDojo.

\begin{table*}
\centering
\caption{Comparison across different controllers using travel time, throughput, delay, accumulated waiting, queue (sec), and CO2 emission (g/sec). For each RL method, we \textbf{bold} the one with the best performance. For all baseline controllers and RL controllers with different feature extractors, we \underline{underline} the method with the best performance. 
} 
\label{tab:comparison-experiment}

\resizebox{\textwidth}{!}{%
\begin{tabular}{cc|cccccc}
\toprule
\multicolumn{2}{c}{} &
  \multicolumn{6}{c}{Two-way single intersection} \\ 

\cmidrule(lr){3-8} 

\multicolumn{2}{c}{Metric} &
  \multicolumn{1}{c}{Travel Time$\hspace{0.1em} \downarrow$ }  &
  \multicolumn{1}{c}{Throughput$\hspace{0.1em}  \uparrow$} &
  \multicolumn{1}{c}{Delay$\hspace{0.1em}  \downarrow$ }  &
  \multicolumn{1}{c}{Accumulated Waiting$\hspace{0.1em}  \downarrow$ } &
  \multicolumn{1}{c}{Queue$\hspace{0.1em}  \downarrow$ } & 
  \multicolumn{1}{c}{CO2 Emissions$\hspace{0.1em}  \downarrow$ }  \\ 
 \midrule 
\multicolumn{2}{c}{FixedTime}&
  \multicolumn{1}{c}{95.11 \std{2.62}} &
  \multicolumn{1}{c}{1879.80 \std{18.74}}&
  \multicolumn{1}{c}{5.57 \std{0.08}} &
  \multicolumn{1}{c}{19480.37 \std{2417.03}} &
  \multicolumn{1}{c}{60.28 \std{2.29}} &
  \multicolumn{1}{c}{190.01 \std{7.11}} \\


\multicolumn{2}{c}{SOTL} &
  \multicolumn{1}{c}{80.51 \std{6.32} } &
  \multicolumn{1}{c}{2429.40 \std{29.58}}&
  \multicolumn{1}{c}{5.88 \std{0.13}} &
  \multicolumn{1}{c}{5009.64 \std{1672.86}} &
  \multicolumn{1}{c}{45.86 \std{6.27} } &
  \multicolumn{1}{c}{151.31 \std{16.53}} \\


\multicolumn{2}{c}{MaxPressure} &
  \multicolumn{1}{c}{44.08 \std{0.89}} &
  \multicolumn{1}{c}{2484.60 \std{2.30}}&
  \multicolumn{1}{c}{4.86 \std{0.10}} &
  \multicolumn{1}{c}{345.40 \std{24.23}} &
  \multicolumn{1}{c}{11.20 \std{0.37}} &
  \multicolumn{1}{c}{56.54 \std{1.27}} \\


\midrule


\multicolumn{2}{c}{DQN-F } &
  \multicolumn{1}{c}{52.35 \std{0.62}} &
  \multicolumn{1}{c}{2478.20 \std{3.42}}&
  \multicolumn{1}{c}{5.02 \std{0.06}} &
  \multicolumn{1}{c}{338.06 \std{10.50}} &
  \multicolumn{1}{c}{15.16 \std{0.30}} &
  \multicolumn{1}{c}{73.92 \std{1.41}} \\

\multicolumn{2}{c}{DQN-MV} &
  \multicolumn{1}{c}{76.13 \std{5.73}} &
  \multicolumn{1}{c}{2225.20 \std{421.93}}&
  \multicolumn{1}{c}{5.73 \std{0.15}} &
  \multicolumn{1}{c}{7646.32 \std{12560.97}} &
  \multicolumn{1}{c}{36.33 \std{10.89}} &
  \multicolumn{1}{c}{141.94 \std{17.47}} \\

\multicolumn{2}{c}{DQN-BEV} &
  \multicolumn{1}{c}{51.01 \std{0.73}} &
  \multicolumn{1}{c}{2475.00 \std{2.00}}&
  \multicolumn{1}{c}{5.07 \std{0.04}} &
  \multicolumn{1}{c}{334.35 \std{6.96}} &
  \multicolumn{1}{c}{14.88 \std{0.24}} &
  \multicolumn{1}{c}{72.93 \std{0.96}} \\

\multicolumn{2}{c}{DQN-BEV*} &
  \multicolumn{1}{c}{\textbf{50.10} \std{0.46}} &
  \multicolumn{1}{c}{\underline{\textbf{2481.40}} \std{3.21}}&
  \multicolumn{1}{c}{\textbf{4.88} \std{0.08}} &
  \multicolumn{1}{c}{\textbf{282.07} \std{13.08}} &
  \multicolumn{1}{c}{\textbf{13.37} \std{0.50}} &
  \multicolumn{1}{c}{\textbf{67.15} \std{1.85}} \\


\midrule


\multicolumn{2}{c}{PPO-F} &
  \multicolumn{1}{c}{\textbf{49.75} \std{0.76}} &
  \multicolumn{1}{c}{2476.40 \std{3.51}}&
  \multicolumn{1}{c}{\textbf{4.95} \std{0.05}} &
  \multicolumn{1}{c}{\textbf{307.79} \std{9.14}} &
  \multicolumn{1}{c}{\textbf{13.54} \std{0.36}}  &
  \multicolumn{1}{c}{\textbf{67.53} \std{1.10}} \\

\multicolumn{2}{c}{PPO-MV} &
  \multicolumn{1}{c}{100.88 \std{0.57}} &
 \multicolumn{1}{c}{2221.00 \std{1.73}} &
\multicolumn{1}{c}{5.73 \std{0.02}} &
\multicolumn{1}{c}{4255.39 \std{24.30}} &
\multicolumn{1}{c}{47.42 \std{0.27}} &
  \multicolumn{1}{c}{171.21 \std{0.92}} \\

\multicolumn{2}{c}{PPO-BEV } &
  \multicolumn{1}{c}{100.01 \std{0.55}} &
 \multicolumn{1}{c}{2120.00 \std{1.41}} &
\multicolumn{1}{c}{5.59 \std{0.02}} &
\multicolumn{1}{c}{4356.62 \std{19.99}} &
\multicolumn{1}{c}{46.59 \std{0.33}} &
  \multicolumn{1}{c}{167.33 \std{1.05}} \\

\multicolumn{2}{c}{PPO-BEV*} &
  \multicolumn{1}{c}{51.00 \std{0.80}} &
 \multicolumn{1}{c}{\textbf{2477.20} \std{3.27}} &
\multicolumn{1}{c}{5.12 \std{0.05}} &
\multicolumn{1}{c}{337.32 \std{7.30}} &
\multicolumn{1}{c}{14.26 \std{0.21}} &
  \multicolumn{1}{c}{70.02 \std{1.03}} \\


\midrule 


\multicolumn{2}{c}{A2C-F} &
  \multicolumn{1}{c}{50.66 \std{1.20}} &
  \multicolumn{1}{c}{2471.40 \std{3.51}}&
  \multicolumn{1}{c}{5.05 \std{0.12}} &
  \multicolumn{1}{c}{340.90 \std{27.45}} &
  \multicolumn{1}{c}{14.85 \std{0.67}} &
  \multicolumn{1}{c}{72.86 \std{2.33}} \\

\multicolumn{2}{c}{A2C-MV} &
 \multicolumn{1}{c}{98.30 \std{0.35}}  &
  \multicolumn{1}{c}{2350.60 \std{4.72}} &
  \multicolumn{1}{c}{5.92 \std{0.01}} &
  \multicolumn{1}{c}{4172.72 \std{24.62}} &
  \multicolumn{1}{c}{48.79 \std{0.10}} &
  \multicolumn{1}{c}{180.39 \std{0.30}} \\

\multicolumn{2}{c}{A2C-BEV} &
 \multicolumn{1}{c}{50.87 \std{0.39}} &
  \multicolumn{1}{c}{2475.40 \std{2.30}} &
  \multicolumn{1}{c}{5.04 \std{0.11}} &
  \multicolumn{1}{c}{324.97 \std{13.33}} &
  \multicolumn{1}{c}{14.58 \std{0.31}} &
  \multicolumn{1}{c}{70.99 \std{1.27}} \\

\multicolumn{2}{c}{A2C-BEV*} &
 \multicolumn{1}{c}{\underline{\textbf{48.67}} \std{0.64}} &
  \multicolumn{1}{c}{\textbf{2480.00} \std{1.58}} &
  \multicolumn{1}{c}{\underline{\textbf{4.87}} \std{0.07}} &
  \multicolumn{1}{c}{\underline{\textbf{277.71}} \std{6.92}} &
  \multicolumn{1}{c}{\underline{\textbf{13.22}} \std{0.22}} &
  \multicolumn{1}{c}{\underline{\textbf{66.35}} \std{1.01}} \\

\midrule
\midrule


\multicolumn{2}{c}{}  &
  \multicolumn{6}{c}{Cologne 1x1} \\

 \cmidrule(lr){3-8} 
\multicolumn{2}{c}{Metric} &
  \multicolumn{1}{c}{Travel Time$\hspace{0.1em} \downarrow$ }  &
  \multicolumn{1}{c}{Throughput$\hspace{0.1em}  \uparrow$} &
  \multicolumn{1}{c}{Delay$\hspace{0.1em}  \downarrow$ }  &
  \multicolumn{1}{c}{Accumulated Waiting$\hspace{0.1em}  \downarrow$ } &
  \multicolumn{1}{c}{Queue$\hspace{0.1em}  \downarrow$ } & 
  \multicolumn{1}{c}{CO2 Emissions$\hspace{0.1em}  \downarrow$ }  \\
 \midrule 
\multicolumn{2}{c}{FixedTime}&
 \multicolumn{1}{c}{165.47 \std{3.33}} &
  \multicolumn{1}{c}{1614.80 \std{34.02}} &
  \multicolumn{1}{c}{4.70 \std{0.02}} &
  \multicolumn{1}{c}{35008.23 \std{2972.30}} &
  \multicolumn{1}{c}{90.85 \std{1.22}} &
  \multicolumn{1}{c}{303.39 \std{1.74}} \\ 

\multicolumn{2}{c}{SOTL} &
  \multicolumn{1}{c}{99.20 \std{6.20}} &
  \multicolumn{1}{c}{1930.20 \std{16.39}} &
  \multicolumn{1}{c}{4.68 \std{0.09}} &
  \multicolumn{1}{c}{5667.02 \std{1179.61}} &
  \multicolumn{1}{c}{32.02 \std{2.13}} &
  \multicolumn{1}{c}{118.32 \std{6.07}} \\ 


\multicolumn{2}{c}{MaxPressure} &
  \multicolumn{1}{c}{44.12 \std{0.40}} &
  \multicolumn{1}{c}{1996.20 \std{2.39}} &
  \multicolumn{1}{c}{2.86 \std{0.08}} &
  \multicolumn{1}{c}{112.05 \std{12.92}} &
  \multicolumn{1}{c}{4.69 \std{0.29}} &
  \multicolumn{1}{c}{46.34 \std{0.82}} \\


\midrule


\multicolumn{2}{c}{DQN-F} &
  \multicolumn{1}{c}{50.28 \std{1.69}} &
  \multicolumn{1}{c}{1994.80 \std{2.17}} &
  \multicolumn{1}{c}{3.12 \std{0.07}} &
  \multicolumn{1}{c}{299.14 \std{19.96}} &
  \multicolumn{1}{c}{6.26 \std{0.26}} &
  \multicolumn{1}{c}{50.77 \std{1.02}} \\

\multicolumn{2}{c}{DQN-MV} &
  \multicolumn{1}{c}{50.63 \std{1.81}} &
  \multicolumn{1}{c}{1981.60 \std{10.19}} &
  \multicolumn{1}{c}{3.41 \std{0.11}} &
  \multicolumn{1}{c}{389.67 \std{90.84}} &
  \multicolumn{1}{c}{8.55 \std{0.76}} &
  \multicolumn{1}{c}{57.89 \std{2.24}} \\

\multicolumn{2}{c}{DQN-BEV} &
  \multicolumn{1}{c}{44.78 \std{1.43}} &
  \multicolumn{1}{c}{1991.60 \std{4.93}} &
  \multicolumn{1}{c}{2.81 \std{0.05}} &
  \multicolumn{1}{c}{114.32 \std{29.87}} &
  \multicolumn{1}{c}{5.14 \std{0.58}} &
  \multicolumn{1}{c}{48.15 \std{1.81}} \\

\multicolumn{2}{c}{DQN-BEV*} &
  \multicolumn{1}{c}{\textbf{43.83} \std{0.72}} &
  \multicolumn{1}{c}{\textbf{1997.60} \std{0.55}} &
  \multicolumn{1}{c}{\textbf{2.74} \std{0.05}} &
  \multicolumn{1}{c}{\textbf{94.78} \std{16.43}} &
  \multicolumn{1}{c}{\textbf{4.39} \std{0.24}} &
  \multicolumn{1}{c}{\textbf{46.27} \std{0.99}} \\


\midrule


\multicolumn{2}{c}{PPO-F} &
  \multicolumn{1}{c}{48.14 \std{0.43}} &
  \multicolumn{1}{c}{\textbf{1997.00} \std{0.00}} &
  \multicolumn{1}{c}{2.97 \std{0.02}} &
  \multicolumn{1}{c}{156.14 \std{17.06}} &
  \multicolumn{1}{c}{5.73 \std{0.18}} &
  \multicolumn{1}{c}{49.72 \std{0.76}} \\

\multicolumn{2}{c}{PPO-MV} &
  \multicolumn{1}{c}{48.44 \std{0.74}} &
  \multicolumn{1}{c}{\textbf{1997.00} \std{0.00}} &
  \multicolumn{1}{c}{3.01 \std{0.03}} &
  \multicolumn{1}{c}{206.09 \std{24.93}} &
  \multicolumn{1}{c}{6.65 \std{0.21}} &
  \multicolumn{1}{c}{52.63 \std{0.64}} \\

\multicolumn{2}{c}{PPO-BEV} &
  \multicolumn{1}{c}{47.71 \std{0.55}} &
  \multicolumn{1}{c}{1996.80 \std{0.45}} &
  \multicolumn{1}{c}{3.00 \std{0.05}} &
  \multicolumn{1}{c}{141.80 \std{15.19}} &
  \multicolumn{1}{c}{5.93 \std{0.19}} &
  \multicolumn{1}{c}{50.62 \std{0.35}} \\

\multicolumn{2}{c}{PPO-BEV*} &
  \multicolumn{1}{c}{\textbf{47.19} \std{0.32}} &
  \multicolumn{1}{c}{\textbf{1997.00} \std{0.00}} &
  \multicolumn{1}{c}{\textbf{2.95} \std{0.03}} &
  \multicolumn{1}{c}{\textbf{122.43} \std{8.22}} &
  \multicolumn{1}{c}{\textbf{5.51} \std{0.12}} &
  \multicolumn{1}{c}{\textbf{49.40} \std{0.55}} \\


\midrule 


\multicolumn{2}{c}{A2C-F} &
  \multicolumn{1}{c}{\underline{\textbf{42.38}} \std{0.42}} &
  \multicolumn{1}{c}{1996.60 \std{0.89}} &
  \multicolumn{1}{c}{2.56 \std{0.01}} &
  \multicolumn{1}{c}{\underline{\textbf{85.57}} \std{8.50}} &
  \multicolumn{1}{c}{3.88 \std{0.02}} &
  \multicolumn{1}{c}{44.72 \std{0.19}} \\

\multicolumn{2}{c}{A2C-MV} &
  \multicolumn{1}{c}{47.64 \std{0.16}} &
  \multicolumn{1}{c}{1997.00 \std{0.71}} &
  \multicolumn{1}{c}{3.03 \std{0.02}} &
  \multicolumn{1}{c}{160.98 \std{14.74}} &
  \multicolumn{1}{c}{6.03 \std{0.17}} &
  \multicolumn{1}{c}{50.65 \std{0.50}} \\

\multicolumn{2}{c}{A2C-BEV} &
  \multicolumn{1}{c}{47.90 \std{0.27}} &
  \multicolumn{1}{c}{1996.80 \std{0.45}} &
  \multicolumn{1}{c}{3.00 \std{0.03}} &
  \multicolumn{1}{c}{184.74 \std{10.40}} &
  \multicolumn{1}{c}{6.17 \std{0.10}} &
  \multicolumn{1}{c}{51.00 \std{0.21}} \\ 

\multicolumn{2}{c}{A2C-BEV*} &
  \multicolumn{1}{c}{43.30 \std{0.60}} &
  \multicolumn{1}{c}{\underline{\textbf{1998.20}} \std{1.30}} &
  \multicolumn{1}{c}{\underline{\textbf{2.55}} \std{0.03}} &
  \multicolumn{1}{c}{89.23 \std{9.51}} &
  \multicolumn{1}{c}{\underline{\textbf{3.71}} \std{0.15}} &
  \multicolumn{1}{c}{\underline{\textbf{43.64}} \std{0.53}} \\

  
\bottomrule
  
\end{tabular}
}
\end{table*}

\begin{figure}[ht]
\centering
\begin{minipage}[b]{0.48\linewidth}
\centering
\includegraphics[width=\textwidth]{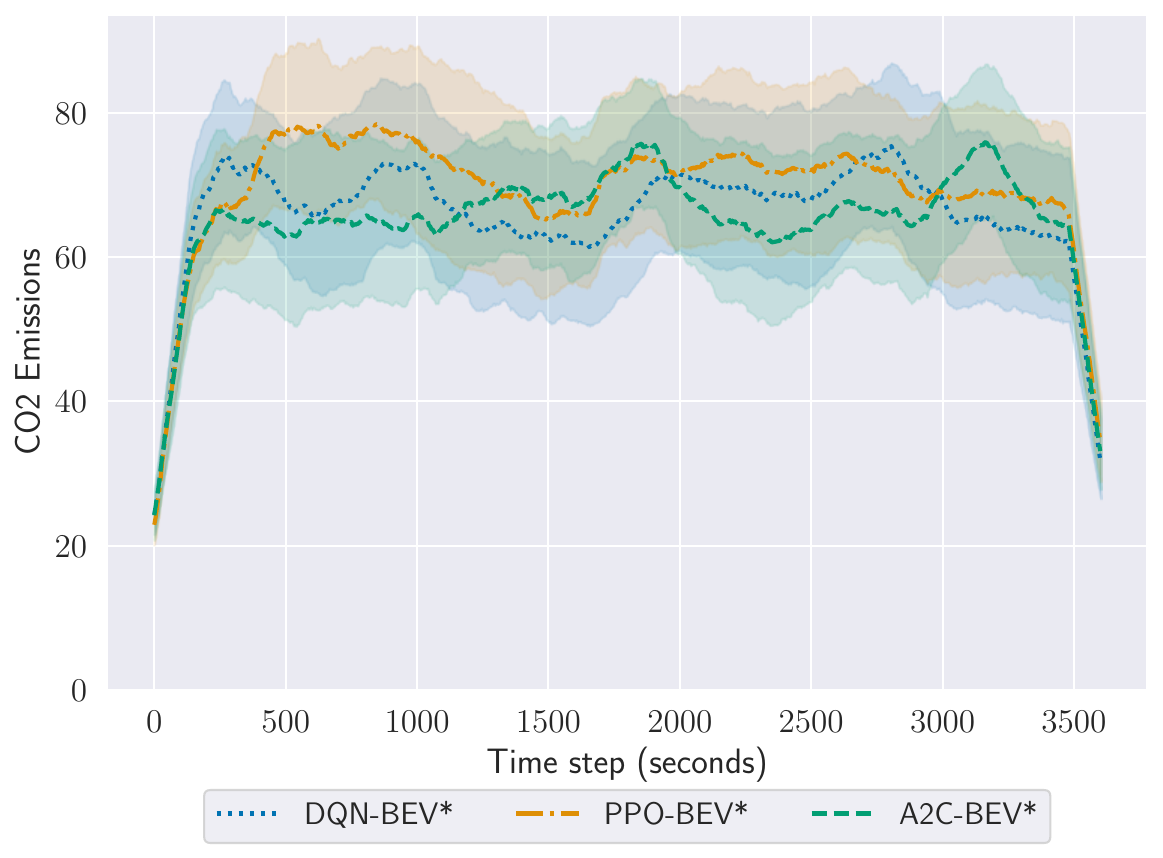}
\end{minipage}
\vspace{-5pt}
\begin{minipage}[b]{0.48\linewidth}
\centering
\includegraphics[width=\textwidth]{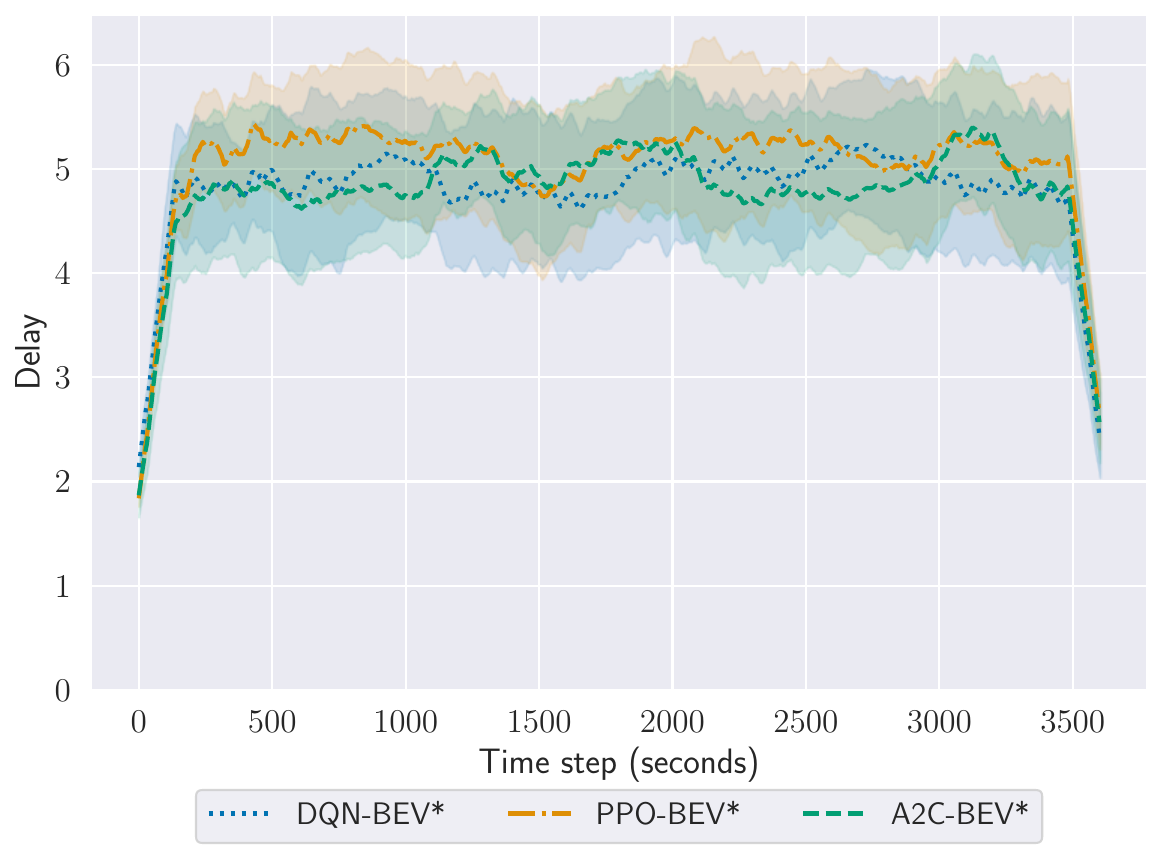}
\end{minipage}
\caption{Evaluation of the trained policies from various RL controllers in the two-way single intersection, with a duration of $3,600$ seconds. The curves are smoothed using a moving average with a window size of $50$.  (\textbf{Left}) Evaluation of CO2 emissions. (\textbf{Right}) Evaluation of average
delay.}
\label{fig:two_way}
\end{figure}

\subsection{Performance Evaluation}

We provide a summary of the experimental results shown in Table~\ref{tab:comparison-experiment}. When trained from scratch, vision-based RL approaches, especially DQN-BEV, PPO-BEV, A2C-BEV, are better or competitive compared to feature-based methods (DQN-F, PPO-F, A2C-F) in terms of performance metrics. However, current feature-based approaches depend on the precise extraction and analysis of aggregated traffic features, such as vehicle density and queue length, which hinders the development of end-to-end solutions. 
In contrast, vision-based approaches excel in end-to-end learning, despite the challenge of directly inferring these aggregated features from images.

More impressively, the use of a frozen foundational model as a feature extractor for vision-based TSC approaches significantly improves the performance of all baseline RL policies. Specifically, in the two-way single intersection scenario, DQN-BEV*, PPO-BEV*, and A2C-BEV* achieve CO2 emissions of $67.15, 70.02$, and $66.35$ g/sec, respectively. This represents a substantial reduction in emissions by $7.93\%, 138.98\%$ and $6.99\%$ compared to the original DQN-BEV, PPO-BEV, and A2C-BEV models. Moreover, PPO-BEV* improves the model convergence of PPO-BEV by providing a robust feature representation; without this enhancement, PPO-BEV would perform poorly. All of these findings highlight the importance of discriminative features in learning superior RL policies, exemplified by the use of foundational models.

We observe a similar trend in the Cologne 1x1 scenario, reducing  emissions from DQN-BEV (48.15), PPO-BEV (50.62) and A2C-BEV (51.00) to DQN-BEV* (46.27), PPO-BEV* (49.40) and A2C-BEV* (43.64), representing a relative reduction of $3.91\%$, $2.41\%$, and $14.43\%$, respectively. We observe consistent improvements in other  metrics by transitioning from feature-based approaches to vision-based approaches.


\begin{figure}[ht]
\centering
\begin{minipage}[b]{0.48\linewidth}
\centering
\includegraphics[width=\textwidth]{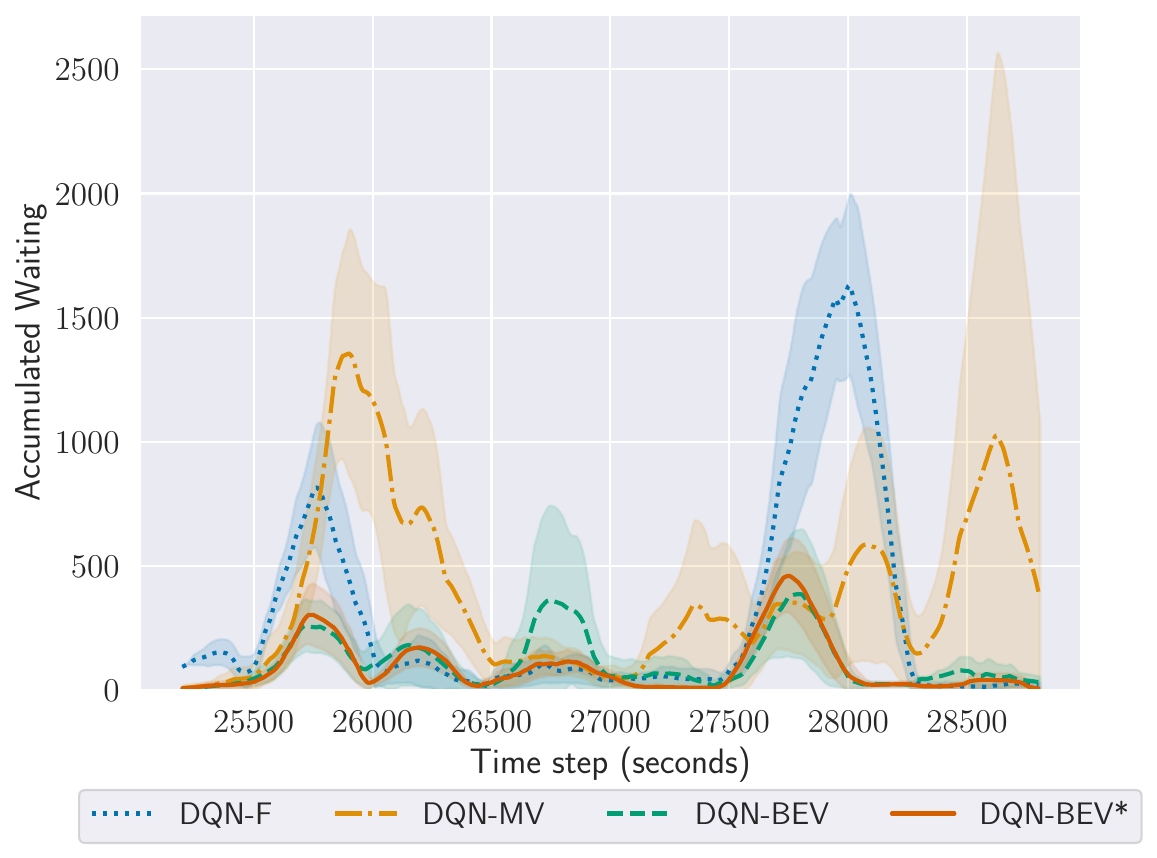}
\end{minipage}
\vspace{-5pt}
\begin{minipage}[b]{0.48\linewidth}
\centering
\includegraphics[width=\textwidth]{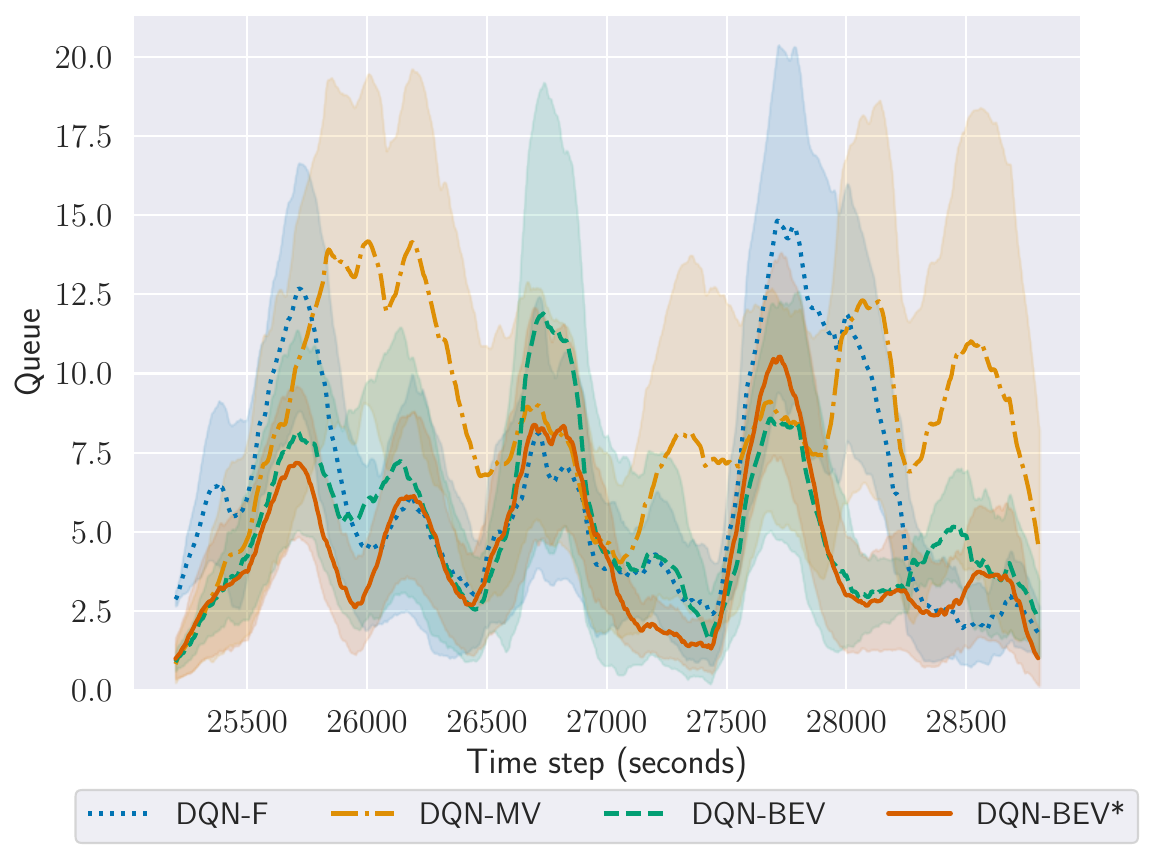}
\end{minipage}
\caption{Evaluation of the trained policies from model variants of DQN in the Cologne 1x1 scenario, with a duration of $3,600$ seconds and a start time of the $25,200_{th}$ second. The curves are smoothed using a moving average with a window size of $50$. (\textbf{Left}) Evaluation of the accumulated waiting time. (\textbf{Right}) Evaluation of average
queue length.}
\label{fig:dqn}
\end{figure}

\subsection{Ablation Study and Visualization}

\underline{Comparison of models with BEV and multi-view camera observations}.  It is highlighted that vision-based RL approaches with a BEV camera demonstrate a clear advantage over multiview cameras. This observation is expected, given that our fusion of features from multiview cameras involves a straightforward concatenation that might ignore useful context information for coordinated control. Additionally, the poor performance of PPO-MV in the two-way single intersection scenario suggests a likely failure in achieving model convergence. We encourage researchers to explore more sophisticated fusion mechanisms to mitigate performance gaps between BEV and multi-view camera observations.

\noindent \underline{Visualizations of different RL controllers}. We further analyzed and plotted the results of CO2 emissions and delay for policies trained by different RL controllers---specifically DQN-BEV*, PPO-BEV*, and A2C-BEV*---at the two-way single intersection. As depicted in Figure~\ref{fig:two_way}, when leveraging discriminative features from foundational models, A2C-BEV* demonstrates slightly superior performance compared to DQN-BEV* and PPO-BEV*. Both DQN-BEV* and A2C-BEV* adopt effective policies that promptly minimize traffic delays and CO2 emissions at critical time points, such as at 0, 900, 1800, and 2700 seconds, where traffic patterns change, leading to superior overall performance.

\noindent  \underline{Visualizations of model variants}. We evaluate the policies learned from various model variants of a single RL algorithm, using DQN as an example. Specifically, we compare DQN-F, DQN-MV, DQN-BEV, and DQN-BEV*. As shown in Figure~\ref{fig:dqn}, DQN-BEV* exhibits negligible fluctuations throughout the evaluation period. All model variants exhibit similar policies at certain timesteps (as evidenced by similar curves and trends), with superior models showing better performance at critical points to minimize sudden increases.



\section{Discussions and Conclusions}

Our current TrafficDojo has several limitations. Firstly, it exclusively supports and evaluates single-agent TSC tasks. However, it is highlighted that our framework is designed with flexibility in mind, and future enhancements are planned to extend its capabilities to accommodate multi-agent TSC tasks where scenarios involve multiple traffic intersections. This incurs additional sensing and computational cost in capturing and processing visual data from the simulated environment. Furthermore, while transitioning from feature-based methods to vision-based methods contributes to a reduction in metrics such as traffic delay and CO2 emissions and facilitates progress towards end-to-end learning, it is imperative to acknowledge that the sim-to-real gap are still observable. Promising directions include works such as~\cite{zhou2024simgen,da2024prompt}.

In this paper,  we have developed a holistic traffic simulation framework towards vision-based TSC with microscopic simulation.
 We have provided a benchmark for training and testing TSC approaches with comprehensive evaluation metrics. Experimental results highlight the promising potential of the end-to-end learning and control for vision-based TSC. It opens up new research opportunities for both reinforcement learning and transportation communities.

\section*{Acknowledgments}

We thank the Panda3D community and the developers of open-source projects such as SUMO-RL and RenderPipeline for their valuable support. This research is supported by the ORAU Ralph E. Powe Junior Faculty Enhancement Award, the NAIRR Pilot Project (NAIRR250417), and Lambda’s Research Grant Program.

\newpage

\bibliographystyle{trb}
\bibliography{trb_template}

@inproceedings{garg2019traffic3d,
  title={Traffic3d: A rich 3d-traffic environment to train intelligent agents},
  author={Garg, Deepeka and Chli, Maria and Vogiatzis, George},
  booktitle={Computational Science--ICCS 2019: 19th International Conference, Faro, Portugal, June 12--14, 2019, Proceedings, Part V 19},
  pages={749--755},
  year={2019},
  organization={Springer}
}

@misc{carla2018release080,
  title        = {CARLA 0.8.0 Release},
  author       = {{CARLA Team}},
  howpublished = {\url{https://carla.org/2018/03/27/release-0.8.0/}},
  note         = {Accessed: 2025-11-20},
  year         = {2018}
}

@inproceedings{pechinger2024sumonity,
  title={Sumonity: Bridging SUMO and Unity for Enhanced Traffic Simulation Experiences},
  author={Pechinger, Mathias and Lindner, Johannes},
  booktitle={SUMO Conference Proceedings},
  volume={5},
  pages={163--177},
  year={2024}
}

@inproceedings{tran2023revisiting,
  title={{Revisiting Pixel-based Traffic Signal Controls using Reinforcement Learning with World Models}},
  author={Tran, Toan V and Sartipi, Mina},
  booktitle={The Workshop on Artificial Intelligence for Social Good at The 37th AAAI conference on artificial intelligence},
  year={2023}
}

@inproceedings{radford2021learning,
  title={Learning transferable visual models from natural language supervision},
  author={Radford, Alec and Kim, Jong Wook and Hallacy, Chris and Ramesh, Aditya and Goh, Gabriel and Agarwal, Sandhini and Sastry, Girish and Askell, Amanda and Mishkin, Pamela and Clark, Jack and others},
  booktitle={International conference on machine learning},
  pages={8748--8763},
  year={2021},
  organization={PMLR}
}

@inproceedings{da2024prompt,
  title={Prompt to Transfer: Sim-to-Real Transfer for Traffic Signal Control with Prompt Learning},
  author={Da, Longchao and Gao, Minquan and Mei, Hao and Wei, Hua},
  booktitle={Proceedings of the AAAI Conference on Artificial Intelligence},
  volume={38},
  number={1},
  pages={82--90},
  year={2024}
}

@article{zhou2024simgen,
  title={SimGen: Simulator-conditioned Driving Scene Generation},
  author={Zhou, Yunsong and Simon, Michael and Peng, Zhenghao and Mo, Sicheng and Zhu, Hongzi and Guo, Minyi and Zhou, Bolei},
  journal={arXiv preprint arXiv:2406.09386},
  year={2024}
}

@software{ilharco_gabriel_2021_5143773,
  author       = {Ilharco, Gabriel and
                  Wortsman, Mitchell and
                  Wightman, Ross and
                  Gordon, Cade and
                  Carlini, Nicholas and
                  Taori, Rohan and
                  Dave, Achal and
                  Shankar, Vaishaal and
                  Namkoong, Hongseok and
                  Miller, John and
                  Hajishirzi, Hannaneh and
                  Farhadi, Ali and
                  Schmidt, Ludwig},
  title        = {OpenCLIP},
  month        = jul,
  year         = 2021,
  note         = {If you use this software, please cite it as below.},
  publisher    = {Zenodo},
  version      = {0.1},
  doi          = {10.5281/zenodo.5143773},
  url          = {https://doi.org/10.5281/zenodo.5143773}
}

@article{wei2019survey,
  title={{A Survey on Traffic Signal Control Methods}},
  author={Wei, Hua and Zheng, Guanjie and Gayah, Vikash and Li, Zhenhui},
  journal={arXiv preprint arXiv:1904.08117},
  year={2019}
}

@inproceedings{pishue2022inrix,
  title={{2022 – INRIX Global Traffic Scorecard}},
  author={Pishue, Bob},
  booktitle={INRIX (December 2021)},
  year={2022}
}

@misc{sumorl,
    author = {Lucas N. Alegre},
    title = {{SUMO-RL}},
    year = {2019},
    publisher = {GitHub},
    journal = {GitHub repository},
    howpublished = {\url{https://github.com/LucasAlegre/sumo-rl}},
}

@inproceedings{lopez2018microscopic,
  title={{Microscopic Traffic Simulation using SUMO}},
  author={Lopez, Pablo Alvarez and Behrisch, Michael and Bieker-Walz, Laura and Erdmann, Jakob and Fl{\"o}tter{\"o}d, Yun-Pang and Hilbrich, Robert and L{\"u}cken, Leonhard and Rummel, Johannes and Wagner, Peter and Wie{\ss}ner, Evamarie},
  booktitle={2018 21st international conference on intelligent transportation systems (ITSC)},
  pages={2575--2582},
  year={2018},
  organization={IEEE}
}

@inproceedings{zhang2019cityflow,
  title={{CityFlow: A Multi-Agent Reinforcement Learning Environment for Large Scale City Traffic Scenario}},
  author={Zhang, Huichu and Feng, Siyuan and Liu, Chang and Ding, Yaoyao and Zhu, Yichen and Zhou, Zihan and Zhang, Weinan and Yu, Yong and Jin, Haiming and Li, Zhenhui},
  booktitle={The world wide web conference},
  pages={3620--3624},
  year={2019}
}

@article{varschen2006mikroskopische,
  title={{Mikroskopische Modellierung der personenverkehrsnachfrage auf Basis von Zeitverwendungstageb{\"u}chern}},
  author={Varschen, Christian and Wagner, Peter},
  journal={Integrierte Mikro-Simulation von Raum-und Verkehrsentwicklung. Theorie, Konzepte, Modelle, Praxis},
  volume={81},
  pages={63--69},
  year={2006},
  publisher={Institut f{\"u}r Stadtbauwesen und Stadtverkehr, RWTH Aachen}
}

@article{schulman2017proximal,
  title={{Proximal Policy Optimization Algorithms}},
  author={Schulman, John and Wolski, Filip and Dhariwal, Prafulla and Radford, Alec and Klimov, Oleg},
  journal={arXiv preprint arXiv:1707.06347},
  year={2017}
}

@article{islam2023effect,
  title={Effect of signal timing on vehicles’ near misses at intersections},
  author={Islam, Zubayer and Abdel-Aty, Mohamed and Goswamy, Amrita and Abdelraouf, Amr and Zheng, Ou},
  journal={Scientific reports},
  volume={13},
  number={1},
  pages={9065},
  year={2023},
  publisher={Nature Publishing Group UK London}
}

@article{ducrocq2023deep,
  title={Deep reinforcement Q-learning for intelligent traffic signal control with partial detection},
  author={Ducrocq, Romain and Farhi, Nadir},
  journal={International journal of intelligent transportation systems research},
  volume={21},
  number={1},
  pages={192--206},
  year={2023},
  publisher={Springer}
}

@misc{mei2022libsignal,
  Author = {Mei, Hao and Lei, Xiaoliang Lei and Da, Longchao and Shi, Bin and Wei, Hua},
  Title = {{LibSignal: An Open Library for Traffic Signal Control}},
  Year = {2022},
  booktitle={NeurIPS Workshop on Reinforcement Learning for Real Life},
}

@article{chen2021distributed,
  title={Distributed dynamic route guidance and signal control for mobile edge computing-enhanced connected vehicle environment},
  author={Chen, Huiyu and Qiu, Tony Z},
  journal={IEEE Transactions on Intelligent Transportation Systems},
  volume={23},
  number={8},
  pages={12251--12262},
  year={2021},
  publisher={IEEE}
}

@phdthesis{song2022impacts,
  title={Impacts of Connected and Autonomous Vehicles on Deep Reinforcement Learning Controlled Intersection Systems},
  author={Song, Li},
  year={2022},
  school={The University of North Carolina at Charlotte}
}

@techreport{fullerton1990traffic,
  title={Traffic Detector Handbook},
  author={Fullerton, Iris J and Kell, James H and Mills, Michael K},
  year={1990},
  institution={United States. Federal Highway Administration}
}

@inproceedings{fellendorf1994vissim,
  title={{VISSIM: A microscopic Simulation Tool to Evaluate Actuated Signal Control including Bus Priority}},
  author={Fellendorf, Martin},
  booktitle={64th Institute of transportation engineers annual meeting},
  volume={32},
  pages={1--9},
  year={1994},
  organization={Springer}
}

@article{hunt1981scoot,
  title={{SCOOT-a Traffic Responsive Method of Coordinating Signals}},
  author={Hunt, PB and Robertson, DI and Bretherton, RD and Winton, RI},
journal={Transport and Road Research Lab., Crowthorne, UK},
  year={1981}
}

@inproceedings{liang2023cblab,
  title={{CBLab: Supporting the Training of Large-scale Traffic Control Policies with Scalable Traffic Simulation}},
  author={Liang, Chumeng and Huang, Zherui and Liu, Yicheng and Liu, Zhanyu and Zheng, Guanjie and Shi, Hanyuan and Wu, Kan and Du, Yuhao and Li, Fuliang and Li, Zhenhui Jessie},
  booktitle={Proceedings of the 29th ACM SIGKDD Conference on Knowledge Discovery and Data Mining},
  pages={4449--4460},
  year={2023}
}

@article{li2022metadrive,
  title={{MetaDrive: Composing Diverse
Driving Scenarios for Generalizable
Reinforcement Learning}},
  author={Li, Quanyi and Peng, Zhenghao and Feng, Lan and Zhang, Qihang and Xue, Zhenghai and Zhou, Bolei},
  journal={IEEE transactions on pattern analysis and machine intelligence},
  volume={45},
  number={3},
  pages={3461--3475},
  year={2022},
  publisher={IEEE}
}

@inproceedings{liang2018rllib,
  title={{RLlib: Abstractions for Distributed Reinforcement Learning}},
  author={Liang, Eric and Liaw, Richard and Nishihara, Robert and Moritz, Philipp and Fox, Roy and Goldberg, Ken and Gonzalez, Joseph and Jordan, Michael and Stoica, Ion},
  booktitle={International conference on machine learning},
  pages={3053--3062},
  year={2018},
  organization={PMLR}
}

@article{raffin2021stable,
  title={{Stable-Baselines3: Reliable Reinforcement Learning Implementations}},
  author={Raffin, Antonin and Hill, Ashley and Gleave, Adam and Kanervisto, Anssi and Ernestus, Maximilian and Dormann, Noah},
  journal={The Journal of Machine Learning Research},
  volume={22},
  number={1},
  pages={12348--12355},
  year={2021},
  publisher={JMLRORG}
}

@inproceedings{dosovitskiy2017carla,
  title={{CARLA: An Open Urban Driving Simulator}},
  author={Dosovitskiy, Alexey and Ros, German and Codevilla, Felipe and Lopez, Antonio and Koltun, Vladlen},
  booktitle={Conference on robot learning},
  pages={1--16},
  year={2017},
  organization={PMLR}
}

@article{miller1963settings,
  title={{Settings for Fixed-Cycle Traffic Signals}},
  author={Miller, Alan J},
  journal={Journal of the Operational Research Society},
  volume={14},
  number={4},
  pages={373--386},
  year={1963},
  publisher={Taylor \& Francis}
}

@article{aslani2017adaptive,
  title={{Adaptive Traffic Signal Control with Actor-critic Methods in a Real-world Traffic Network with Different Traffic Disruption Events}},
  author={Aslani, Mohammad and Mesgari, Mohammad Saadi and Wiering, Marco},
  journal={Transportation Research Part C: Emerging Technologies},
  volume={85},
  pages={732--752},
  year={2017},
  publisher={Elsevier}
}

@article{casas2017deep,
  title={{Deep Deterministic Policy Gradient for Urban Traffic Light Control}},
  author={Casas, Noe},
  journal={arXiv preprint arXiv:1703.09035},
  year={2017}
}

@article{mannion2016experimental,
  title={{An Experimental Review of Reinforcement Learning Algorithms for Adaptive Traffic Signal Control}},
  author={Mannion, Patrick and Duggan, Jim and Howley, Enda},
  journal={Autonomic road transport support systems},
  pages={47--66},
  year={2016},
  publisher={Springer}
}

@inproceedings{pham2013learning,
  title={{Learning Coordinated Traffic Light Control}},
  author={Pham, Tong Thanh and Brys, Tim and Taylor, Matthew E and Brys, Tim and Drugan, Madalina M and Bosman, PA and Cock, Martine-De and Lazar, Cosmin and Demarchi, L and Steenhoff, David and others},
  booktitle={Proceedings of the Adaptive and Learning Agents workshop (at AAMAS-13)},
  volume={10},
  pages={1196--1201},
  year={2013},
  organization={IEEE}
}

@article{van2016coordinated,
  title={{Coordinated Deep Reinforcement Learners
for Traffic Light Control}},
  author={Van der Pol, Elise and Oliehoek, Frans A},
  journal={Proceedings of learning, inference and control of multi-agent systems (at NIPS 2016)},
  volume={8},
  pages={21--38},
  year={2016}
}

@article{arel2010reinforcement,
  title={Reinforcement Learning-based Multi-agent System for Network Traffic Signal Control},
  author={Arel, Itamar and Liu, Cong and Urbanik, Tom and Kohls, Airton G},
  journal={IET Intelligent Transport Systems},
  volume={4},
  number={2},
  pages={128--135},
  year={2010},
  publisher={IET}
}

@article{zheng2019diagnosing,
  title={{Diagnosing Reinforcement Learning for Traffic Signal Control}},
  author={Zheng, Guanjie and Zang, Xinshi and Xu, Nan and Wei, Hua and Yu, Zhengyao and Gayah, Vikash and Xu, Kai and Li, Zhenhui},
  journal={arXiv preprint arXiv:1905.04716},
  year={2019}
}

@article{arulkumaran2017brief,
  title={{A Brief Survey of Deep Reinforcement Learning}},
  author={Arulkumaran, Kai and Deisenroth, Marc Peter and Brundage, Miles and Bharath, Anil Anthony},
  journal={arXiv preprint arXiv:1708.05866},
  year={2017}
}

@article{boxill2000evaluation,
  title={{An Evaluation of Traffic Simulation Models for
Supporting ITS Development}},
  author={Boxill, Sharon Adams and Yu, Lei},
  journal={Houston, TX: Development Centre for Transportation Training and Research, Texas Southern University},
  year={2000}
}

@article{wu2017flow,
  title={{Flow: Architecture and Benchmarking for
Reinforcement Learning in Traffic Control}},
  author={Wu, Cathy and Kreidieh, Aboudy and Parvate, Kanaad and Vinitsky, Eugene and Bayen, Alexandre M},
  journal={arXiv preprint arXiv:1710.05465},
  volume={10},
  year={2017}
}

@inproceedings{parish2001procedural,
  title={{Procedural Modeling of Cities}},
  author={Parish, Yoav IH and M{\"u}ller, Pascal},
  booktitle={Proceedings of the 28th annual conference on Computer graphics and interactive techniques},
  pages={301--308},
  year={2001}
}

@inproceedings{dupuis2010opendrive,
  title={{OpenDRIVE 2010 and Beyond – Status and Future of the de facto Standard for the Description of Road Networks}},
  author={Dupuis, Marius and Strobl, Martin and Grezlikowski, Hans},
  booktitle={Proc. of the Driving Simulation Conference Europe},
  pages={231--242},
  year={2010}
}

@article{ma2021traffic,
  title={{Traffic Scenarios for Automated Vehicle Testing: A Review of Description Languages and Systems}},
  author={Ma, Jing and Che, Xiaobo and Li, Yanqiang and Lai, Edmund M-K},
  journal={Machines},
  volume={9},
  number={12},
  pages={342},
  year={2021},
  publisher={MDPI}
}

@misc{asam2023openCRG,
  author={ASAM},
  title = {{OpenCRG}},
  howpublished = "\url{https://www.asam.net/standards/detail/opencrg/}",
  year = {2023}, 
  note = "[Online; accessed 1-December-2023]"
}

@inproceedings{althoff2017commonroad,
  title={{CommonRoad: Composable Benchmarks for Motion Planning on Roads}},
  author={Althoff, Matthias and Koschi, Markus and Manzinger, Stefanie},
  booktitle={2017 IEEE Intelligent Vehicles Symposium (IV)},
  pages={719--726},
  year={2017},
  organization={IEEE}
}

@article{mnih2015human,
  title={{Human-Level Control Through Deep Reinforcement Learning}},
  author={Mnih, Volodymyr and Kavukcuoglu, Koray and Silver, David and Rusu, Andrei A and Veness, Joel and Bellemare, Marc G and Graves, Alex and Riedmiller, Martin and Fidjeland, Andreas K and Ostrovski, Georg and others},
  journal={nature},
  volume={518},
  number={7540},
  pages={529--533},
  year={2015},
  publisher={Nature Publishing Group}
}

@article{cools2013self,
  title={{Self-Organizing Traffic Lights: A Realistic Simulation}},
  author={Cools, Seung-Bae and Gershenson, Carlos and D’Hooghe, Bart},
  journal={Advances in applied self-organizing systems},
  pages={45--55},
  year={2013},
  publisher={Springer}
}

@incollection{varaiya2013max,
  title={{The Max-Pressure Controller for Arbitrary Networks of Signalized Intersections}},
  author={Varaiya, Pravin},
  booktitle={Advances in dynamic network modeling in complex transportation systems},
  pages={27--66},
  year={2013},
  publisher={Springer}
}

@inproceedings{mnih2016asynchronous,
  title={{Asynchronous Methods for Deep Reinforcement Learning}},
  author={Mnih, Volodymyr and Badia, Adria Puigdomenech and Mirza, Mehdi and Graves, Alex and Lillicrap, Timothy and Harley, Tim and Silver, David and Kavukcuoglu, Koray},
  booktitle={International conference on machine learning},
  pages={1928--1937},
  year={2016},
  organization={PMLR}
}

@inproceedings{ault2021reinforcement,
  title={{Reinforcement Learning Benchmarks
for Traffic Signal Control}},
  author={Ault, James and Sharon, Guni},
  booktitle={Thirty-fifth Conference on Neural Information Processing Systems Datasets and Benchmarks Track (Round 1)},
  year={2021}
}

@inproceedings{bender2014lanelets,
  title={{ Lanelets: Efficient Map Representation for Autonomous Driving}},
  author={Bender, Philipp and Ziegler, Julius and Stiller, Christoph},
  booktitle={2014 IEEE Intelligent Vehicles Symposium Proceedings},
  pages={420--425},
  year={2014},
  organization={IEEE}
}

@inproceedings{fremont2019scenic,
  title={{Scenic: A Language for Scenario Specification and Scene Generation}},
  author={Fremont, Daniel J and Dreossi, Tommaso and Ghosh, Shromona and Yue, Xiangyu and Sangiovanni-Vincentelli, Alberto L and Seshia, Sanjit A},
  booktitle={Proceedings of the 40th ACM SIGPLAN Conference on Programming Language Design and Implementation},
  pages={63--78},
  year={2019}
}
\end{document}